\documentclass[letterpaper, 10 pt, conference]{ieeeconf}
\usepackage{graphicx} %
\usepackage{amsmath}
\usepackage{amssymb}
\usepackage{booktabs}
\usepackage{array}
\usepackage{algorithm}
\usepackage{algpseudocode}
\usepackage{xcolor}
\usepackage{censor}
\usepackage[colorlinks=true, linkcolor=black, citecolor=black, urlcolor=blue]{hyperref}

\usepackage{float}

\IEEEoverridecommandlockouts 
\title{\LARGE \bf
SMAT: Staged Multi-Agent Training for Co-Adaptive Exoskeleton Control
}

\author{Yifei Yuan, Ghaith Androwis, and Xianlian Zhou%
\thanks{Yifei Yuan and Xianlian Zhou are with the Department of Biomedical Engineering, New Jersey Institute of Technology, Newark, NJ 07102, USA (yy72@njit.edu, alexzhou@njit.edu). Ghaith Androwis (GAndrowis@kesslerfoundation.org) is with Kessler Foundation, West Orange, NJ, 07052, USA. This work was supported by NSF Award \#2524089 and NIDILRR funded RERC Grant 90REGE0025-01-00.}
}

\begin{document}
\maketitle
\thispagestyle{empty}
\pagestyle{empty}

\begin{abstract}
Effective exoskeleton assistance requires co-adaptation: as the device alters joint dynamics, the user reorganizes neuromuscular coordination, creating a non-stationary learning problem. Most learning-based approaches do not explicitly account for the sequential nature of human motor adaptation, leading to training instability and poorly timed assistance. We propose Staged Multi-Agent Training (SMAT), a four-stage curriculum inspired by how users acclimate to a wearable device. In SMAT, a musculoskeletal human actor and a bilateral hip exoskeleton actor are trained progressively: the human first learns unassisted gait, then adapts to the added device mass; the exoskeleton subsequently learns a positive assistance pattern against a stabilized human policy, and finally both agents co-adapt with full torque capacity and bidirectional feedback. We implement SMAT in the MyoAssist simulation environment using a 26-muscle lower-limb model and an attached hip exoskeleton. Our musculoskeletal simulations demonstrate that the learned exoskeleton control policy produces an average \textbf{10.1\%} reduction in simulated hip muscle activation relative to the no-assist condition. We validated the learned controller in an offline setting using open-source gait data, then deployed it to a physical hip exoskeleton for treadmill experiments with five subjects. The resulting policy delivers consistent assistance and predominantly positive mechanical power without the need for any explicitly imposed timing shift (mean positive power: 13.6\,W at 6\,Nm RMS torque to 23.8\,W at 9.3\,Nm RMS torque, with minimal negative power) consistently across all subjects without subject-specific retraining.
\end{abstract}

\section{Introduction}

Lower-limb exoskeletons offer a promising avenue for gait rehabilitation and physical augmentation, but effective assistance requires controllers that can adapt to the unique biomechanics of each user~\cite{rodriguez2021systematic,baud2021review}. In practice, this is a co-adaptation problem: as the device adjusts assistance, the user simultaneously reorganizes neuromuscular coordination, altering joint kinematics, muscle activations, and inter-limb timing~\cite{poggensee2021adaptation}. The result is a tightly coupled human–device system in which both agents influence each other.

Recent advances in AI and reinforcement learning (RL) have enabled data-driven optimization of exoskeleton control policies in simulation. 
Musculoskeletal simulation combined with RL has demonstrated robust policy learning for exoskeleton control~\cite{luo2023robust}, and joint human-exoskeleton training with sim-to-real transfer has been demonstrated for walking and related activities~\cite{luo2024experiment}. %
Subsequent studies extended simulation-based RL to broader assistance scenarios, including human-aligned frameworks that predict adaptation across walking speeds~\cite{leem2026exo}, ankle assistance for gait asymmetry~\cite{yuan2026gait}, and multi-joint squatting tasks~\cite{ratnakumar2026reinforcement}. Yet none of these approaches explicitly accounts for how human motor coordination evolves as users acclimate to the device. 

Despite these advances, a fundamental challenge remains unresolved: the learning problem is inherently non-stationary. As the exoskeleton policy improves, the human policy simultaneously adapts, altering the underlying state distribution. Without structured coordination, jointly optimizing both agents can lead to instability, including oscillatory torque outputs, poorly timed assistance, and destabilized training dynamics. Existing approaches typically address reward design or policy robustness, but they do not explicitly model or structure the sequential nature of human motor adaptation during device acclimation.

Prior work on musculoskeletal skill learning suggests that curriculum-based reinforcement learning can improve training stability and the physiological plausibility of learned motor policies~\cite{bengio2009curriculum,chiappa2024acquiring,caggiano2022myosuite}. Analogously, human users often need to stabilize gait coordination before they can reliably benefit from active assistance. These observations motivate a structured training protocol that treats motor adaptation as a staged process, which progressively introduces the device load and assistance as the human policy stabilizes.

To address this gap, we present \textbf{SMAT} (Staged Multi-Agent Training), a structured curriculum learning protocol for co-adaptive hip exoskeleton control. SMAT trains two control policies (human policy $\pi_h$ and exoskeleton policy $\pi_e$) through four stages: (1) baseline gait learning, (2) adaptation to the attached exoskeleton mass, (3) exoskeleton assistance pattern learning with a frozen human policy, and (4) final human-exoskeleton co-adaptation. By progressively introducing device load and assistance while stabilizing the human policy, SMAT reduces non-stationarity and improves training robustness.

\begin{figure*}[t]
    \centering
    \includegraphics[width=\textwidth]{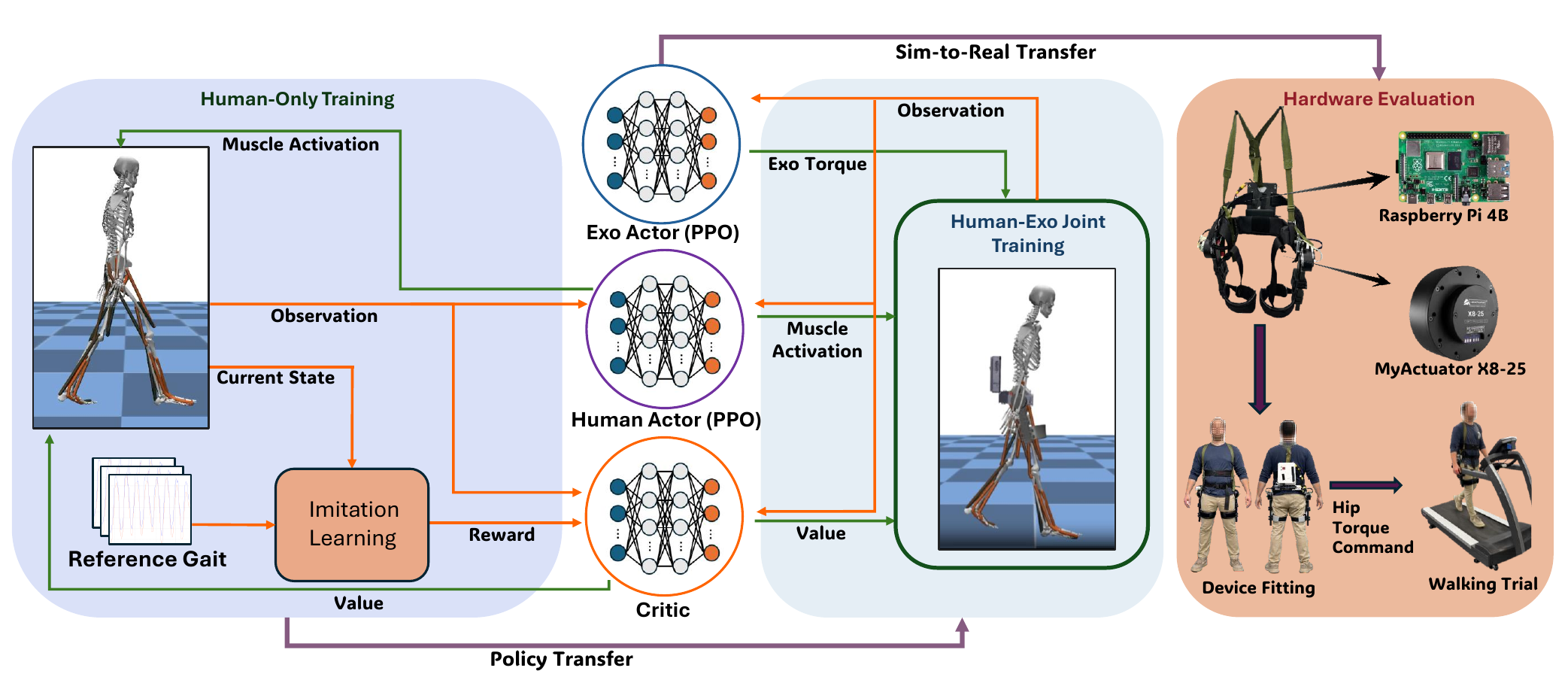}
    \caption{Overview of the SMAT framework. Two PPO-based actors, a 
        musculoskeletal human actor $\pi_h$ and an exoskeleton actor $\pi_e$, interact 
        with a shared physics simulation through a shared critic $V_\psi$. 
        \textit{Human-Only Training} (left): only $\pi_h$ updates via imitation 
        learning from a reference gait. \textit{Human-Exo Joint Training} (center): 
        both actors train simultaneously. \textit{Hardware Evaluation} (right): the 
        trained exoskeleton policy is deployed via sim-to-real transfer, running on 
        a Raspberry Pi 4B to command bilateral MyActuator X8-25 actuators during 
        treadmill walking trials. In SMAT, Stage 1 resides in the human only training mode; Stage 2 adapts the human actor with added exoskeleton mass; Stage 3 freezes the human actor and only trains the exo actor for positive assistance timing; and finally Stage 4 trains both actors for co-adaptation. Policies trained in each stage are subsequently transferred to the next stage.}
    \label{fig:framework}
    \vspace{-0.05in}
\end{figure*}

We implement SMAT in the open source MyoAssist simulation environment~\cite{tan2025myoassist} using a 26-muscle lower-limb model. The learned controller is evaluated using open-source gait data and online treadmill experiments with five human subjects. The main contributions of this work are: (1) a staged multi-agent training framework for co-adaptive hip exoskeleton control that improves training robustness through stage-wise reward decomposition, (2) a modular four-stage training pipeline that separates human adaptation and assistance learning and is in principle extensible to other assistive devices, (3) simulation and hardware validation demonstrating reduced hip muscle activation in simulation, smooth and stable torque profiles, and potent assistive power across subjects; and (4) an ablation analysis of training instabilities in co-adaptive RL for assistive devices. 
The code is publicly available at \url{https://github.com/Fitz-Yuan/SMAT-Exo}.

\section{Method}

\subsection{Multi-Agent Reinforcement Learning Framework}
\label{subsec:framework}

We formulate hip exoskeleton assistance as a multi-agent reinforcement learning (MARL)
problem~\cite{luo2024experiment,tan2025myoassist} and implement it within a configurable actor-critic framework optimized with proximal policy optimization (PPO)~\cite{schulman2017proximal}. From MyoAssist~\cite{tan2025myoassist} we inherit the musculoskeletal model, imitation reward terms, and PPO actor--critic architecture. The staged curriculum, the exoskeleton actor and its observation and reward design, and the mid-training observation augmentation and freezing schedule are new to this work. As illustrated in Fig.~\ref{fig:framework}, the framework includes two learning
agents, a human actor $\pi_h$ and an exoskeleton actor $\pi_e$, operating in a
shared physics simulation that couples a musculoskeletal human model with an
exoskeleton.

The human actor receives a 65-dimensional observation (16 joint angles, 18 joint velocities, 26 muscle activations, 4 foot-contact forces, and the target speed) and directly
outputs 26 muscle activations $\mathbf{a}\in[0,1]^{26}$ (13 per side), implemented as an MLP with hidden sizes [256, 128]. The policy directly controls muscle excitations, allowing emergent gait patterns to arise from neuromuscular dynamics.
The exoskeleton actor receives an 18-dimensional observation $\mathbf{o}_e$
consisting of a three-step history of bilateral hip flexion angles and
angular velocities (12 dimensions, 4 dims per timestep) together with a three-step
history of its own action (normalized torque) outputs $[\hat{u}_r, \hat{u}_l]$
(6 dimensions, 2 dims per timestep).
The exoskeleton actor outputs normalized bilateral hip torque
commands $\hat{\mathbf{u}}\in[-1,1]^{2}$, which are scaled by stage-dependent torque limits to produce physical assistance, implemented as an MLP with hidden sizes [128, 64]. Both actors share a critic with hidden sizes [256, 128].

\subsection{Multi-stage Curriculum Training}

\begin{table}[t]
\centering
\caption{Stage-wise reward configuration and weights used in the multi-stage curriculum.}
\label{tab:reward_weights}
\begin{tabular*}{\columnwidth}{@{\extracolsep{\fill}}lcccc}
\toprule
Reward term & Stage 1 & Stage 2 & Stage 3 & Stage 4 \\
\midrule
$r_{\text{fwd}}$                   & 0.8   & 0.8   & 1.5  & 1.5 \\
$r_{\text{muscle}}$                & 0.01  & 0.01  & 0.15 & 0.15 \\
$r_{\Delta a}$                     & 0.005 & 0.005 & 0.05 & 0.05 \\
$r_{\text{hip-act}}$               & \textemdash & \textemdash & 2.0 & 5.0 \\
$r_{\text{exo}}$                   & \textemdash & \textemdash & 4.0 & 4.0 \\
$r_{\Delta\tau}$                   & \textemdash & \textemdash & \textemdash & 1.0 \\
Imitation ($r_{\text{qpos/qvel}}$) & 1.0    & 1.0    & 1.0*  & 1.0* \\
$r_{\text{constraint}}$            & \textemdash & \textemdash & 0.5 & 0.5 \\
$r_{\text{foot}}$                  & \textemdash & \textemdash & 0.3 & 0.3 \\
\bottomrule
\end{tabular*}

\vspace{0.5mm}
\begin{minipage}{\columnwidth}
\footnotesize
* Hip flexion imitation weights are set to zero in Stages 3 and 4 while other joint imitation terms remain active. \textemdash\ indicates that the term is inactive in that stage. In this table, $r_{\text{exo}}$ denotes the Stage 3 exoskeleton reward in Eq.~(\ref{eq:rexostage3}) and the Stage 4 exoskeleton reward in Eq.~(\ref{eq:rexoreward_stage4}).
\end{minipage}
\end{table}

To mitigate non-stationarity, we decompose training into four stages, each isolating one adaptation challenge while progressively introducing mechanical coupling and active assistance. Algorithm~\ref{alg:multistage} summarizes the schedule, freezing, and reward activation.

The stages are designed to sequentially address:
(1) Stable human gait generation,
(2) Adaptation to exoskeleton mass and inertia,
(3) Learning assistance timing,
(4) Full human–exoskeleton co-adaptation.
This progression avoids optimizing all objectives at once, improving stability and convergence.

\textbf{Stage 1 (Human Baseline Gait Learning).}
Stage 1 trains a human-only locomotion policy without the exoskeleton.
The goal is to obtain a stable walking policy that mimics a reference
gait and serves as the baseline for later stages. The reference gait is the 1.25\,m/s walking trajectory provided with MyoAssist~\cite{tan2025myoassist}, which sets the nominal training speed.
The per-timestep reward, with imitation terms adopted from~\cite{tan2025myoassist}, is

\begin{equation}
\begin{split}
r^{(1)} = \;&w_{\text{fwd}}\,r_{\text{fwd}}
           + w_{\text{mus}}\,r_{\text{muscle}}\\
           &+ w_{\text{qpos}}\,r_{\text{qpos}}
           + w_{\text{qvel}}\,r_{\text{qvel}}
           + w_{\Delta a}\,r_{\Delta a},
\end{split}
\label{eq:reward_stage1}
\end{equation}

where
\begin{subequations}
\label{eq:reward_terms}
\begin{align}
r_{\text{fwd}}     &= \Delta t \cdot
  \exp\!\left(-5\,(v - v^*)^2\right),
  \label{eq:rfwd}\\
r_{\text{muscle}}  &= -\frac{\Delta t}{N_m}
  \sum_{j=1}^{N_m} a_j,
  \label{eq:rmuscle}\\
r_{\text{qpos}}    &= \Delta t
  \sum_{i\in\mathcal{J}} w_i\,
  \exp\!\left(-8\,(q_i - q_i^{\text{ref}})^2\right),
  \label{eq:rqpos}\\
r_{\text{qvel}}    &= \Delta t
  \sum_{i\in\mathcal{J}} w_i\,
  \exp\!\left(-8\,(\dot{q}_i - \rho\,\dot{q}_i^{\text{ref}})^2\right),
  \label{eq:rqvel}\\
r_{\Delta a}       &= \frac{\Delta t}{N_m}
  \sum_{j=1}^{N_m}
  \exp\!\left(-4\,(a_{j,t}-a_{j,t-1})^2\right).
  \label{eq:rdeltaa}
\end{align}
\end{subequations}
Here $v$ is the forward pelvis speed, $v^*{=}1.25$ m/s is the target speed,
$N_m{=}26$ is the total number of muscles, $a_{j,t}\!\in\![0,1]$ is the
activation of muscle $j$ at timestep $t$ with $a_{j,t-1}$ denoting the previous-step activation,
$q_i$ and $\dot{q}_i$ denote the angle and angular velocity of joint $i$, with reference values $q_i^{\text{ref}}$ and $\dot{q}_i^{\text{ref}}$. Here $\mathcal{J}$ is the set of tracked joints with
per-joint imitation weights $w_i$, and
$\rho = v^*/\dot{q}_{\text{pelvis}}^{\text{ref}}(t)$, where $\dot{q}_{\text{pelvis}}^{\text{ref}}$ is the reference forward pelvis velocity,
scales the reference joint velocities to match the target walking speed.
The control timestep is $\Delta t{=}0.02$\,s and multiplying each term by $\Delta t$
makes the reward independent of control frequency.

\begin{algorithm}[t]
\caption{Multi-stage Curriculum Training for Human--Exoskeleton Co-adaptation}
\label{alg:multistage}
\begin{algorithmic}[1]
\Require $v^\star=1.25$ m/s, $\Delta t=0.02$ s (50 Hz), $\gamma=0.99$
\Ensure Human policy $\pi_h$ and exoskeleton policy $\pi_e$
\State Per-stage simulation steps: $T_1, T_2, T_3, T_4$
\State Initialize $\pi_h,\pi_e$ and shared critic $V_\psi$

\State \textbf{Stage 1: Human baseline gait (imitation, human-only)}
\For{$t=1$ to $T_1$}
  \State Roll out with $\pi_h$; update $(\pi_h,V_\psi)$ using PPO~\cite{schulman2017proximal}
\EndFor
\State Save $\pi_h^{(1)}$

\State \textbf{Stage 2: Human adapts to a passive exoskeleton}
\State Attach exoskeleton structure (mass/inertia added); load $\pi_h^{(1)}$; set exoskeleton torque to zero
\For{$t=1$ to $T_2$}
  \State Roll out with $\pi_h$; update $(\pi_h,V_\psi)$ using PPO
\EndFor
\State Save $\pi_h^{(2)}$

\State \textbf{Stage 3: Exoskeleton pattern learning (human frozen)}
\State Set $\tau_{\max}\!\leftarrow\!6$ Nm; freeze $\pi_h^{(2)}$; re-initialize $\pi_e$
\State Disable hip imitation terms; add $r_{\text{hip-act}}$ (Eq.~\eqref{eq:rhipact}) and $r_{\text{exo}}^{(3)}$ (Eq.~\eqref{eq:rexostage3})
\For{$t=1$ to $T_3$}
  \State Roll out with frozen $\pi_h$ and acting $\pi_e$; update $(\pi_e,V_\psi)$ using PPO
\EndFor
\State Save $\pi_e^{(3)}$

\State \textbf{Stage 4: Co-adaptation with power-based assistance}
\State Set $\tau_{\max}\!\leftarrow\!25$ Nm; load $\pi_e^{(3)}$; unfreeze $\pi_h^{(2)}$
\State Augment human observation with current exo torques $[\hat{u}_r,\hat{u}_l]$
\State Expand human actor input by 2 dims; copy existing input weights; randomly initialize weights for the 2 new input dims
\State Update $r_{\text{exo}}^{(4)}$ (Eq.~\eqref{eq:rexoreward_stage4}) and add $r_{\Delta\tau}$ (Eq.~\eqref{eq:ractionrate})
\For{$t=1$ to $T_4$}
  \State Roll out with $(\pi_h,\pi_e)$; update $(\pi_h,\pi_e,V_\psi)$ using PPO
\EndFor
\end{algorithmic}
\end{algorithm}

\textbf{Stage 2 (Adaptation to Added Exoskeleton Mass).}
Stage 2 starts from the Stage 1 policy and attaches the exoskeleton to the musculoskeletal model. The pelvis frame and bilateral thigh links are parented to the corresponding body segments, coupling device mass and inertia into the body dynamics. The actuator is an ideal torque source acting coaxially at the human hip flexion joints, with passive abduction joints providing lateral compliance, while the pelvis and thigh attachments use compliant elements emulating strap deformation. In this stage, the exoskeleton actuator torque limit is set to zero in the simulation model, so no assistance torque is applied regardless of the actor output (Algorithm~\ref{alg:multistage}). The human policy continues training with the same reward in Eqs.~\eqref{eq:reward_stage1} and \eqref{eq:reward_terms}. This stage isolates human adaptation to the added exoskeleton mass and inertia before active exoskeleton assistance is introduced.

\textbf{Stage 3 (Learning of Assistance Timing with Frozen Human Policy).}
Stage 3 freezes the human policy and trains only the exoskeleton policy, using the stable Stage~2 gait as a fixed environment for initial timing learning. Hip imitation terms are disabled so the exoskeleton cannot exploit tracking rewards by resisting hip motion instead of assisting the hip flexor and extensor groups.

A hip muscle activation term is added in Stage 3:
\begin{equation}
r_{\text{hip-act}}
=
-\Delta t \cdot \frac{1}{|\mathcal{M}_{\text{hip}}|}
\sum_{m \in \mathcal{M}_{\text{hip}}}
a_m^2,
\label{eq:rhipact}
\end{equation}
where $a_m \in [0,1]$ is the activation of muscle $m$, and $\mathcal{M}_{\text{hip}}$ includes iliopsoas, gluteus maximus, rectus femoris, and hamstrings on both legs.

In Stage 3, the exoskeleton torque limit is set to $\tau_{\max}=6$\,Nm to constrain exploration during initial pattern learning and limit its perturbation to human control. The following reward is used to encourage positive assistance based on power sign and the use of higher assistance torques:
\begin{equation}
r_{\text{exo}}^{(3)}
=
\Delta t \sum_{j\in\{l,r\}}
\alpha_d\,\hat{u}_j^2\,
\mathrm{sign}(\hat{u}_j \omega_j),
\label{eq:rexostage3}
\end{equation}
where $\hat{u}_j\in[-1,1]$ is the normalized torque command output by the exoskeleton actor and $\omega_j$ is the hip joint angular velocity. We set $\alpha_d = 0.5$.
The Stage~3 exoskeleton policy is used to initialize Stage~4 joint training.

\textbf{Stage 4 (Human-Exoskeleton Co-Adaptation).}

Stage 4 starts the joint training of the human and exoskeleton policies for co-adaptation. The exoskeleton policy is loaded from Stage 3, the human policy is unfrozen, and the exoskeleton torque limit is set to the maximum motor torque $\tau_{\max}=25$ Nm (Algorithm~\ref{alg:multistage}). The human actor observation is augmented with the current normalized exoskeleton torques $[\hat{u}_r,\hat{u}_l]$. In implementation, we expand the human actor input by two dimensions, keep the learned Stage 2 human actor weights for the original input dimensions, and randomly initialize the weights associated with the two new input dimensions. This keeps the learned walking policy while allowing the human actor to learn responses to exoskeleton torque during co-adaptation.

Both actors are trained jointly with a shared critic using the full system state, optimizing a single shared team reward, the sum of the active terms in Table~\ref{tab:reward_weights}, rather than agent-specific rewards. Stage 4 replaces the Stage 3 positive timing assistance reward ($r_{\text{exo}}^{(3)}$) with a power- and smoothness-based reward ($r_{\text{exo}}^{(4)}$), as the Stage~3 reward ($r_{\text{exo}}^{(3)}$) tends to produce saturated exoskeleton torques:
\begin{equation}
r_{\text{exo}}^{(4)}
=
\Delta t
\sum_{j \in \{l,r\}}
\left(
\alpha \hat{u}_j \hat{\omega}_j
-
\beta \hat{u}_j^2
-
\lambda_s \max\!\left(0, |\hat{u}_j|-\delta\right)^2
\right),
\label{eq:rexoreward_stage4}
\end{equation}
where $\hat{u}_j\in[-1,1]$ is the normalized torque command and $\hat{\omega}_j=\mathrm{clip}(\omega_j/\omega_s,-1,1)$ is the normalized joint angular velocity (with $\omega_s=2.0~\mathrm{rad/s}$). The first term is related to mechanical power, rewarding torque that assists joint motion and penalizing torque that resists it. The second term penalizes torque magnitude to prevent the policy from preferring large torques. The third term adds a stronger penalty above the threshold $\delta$, preventing actuator saturation when other reward terms drive torque upward. We set $\alpha=0.3$, $\beta=0.15$, $\lambda_s=2.0$, and $\delta=0.8$.

An exoskeleton torque-rate penalty is also added in Stage 4:
\begin{equation}
r_{\Delta\tau}
=
-\Delta t
\sum_{j \in \{l,r\}}
\left(\hat{u}_{j,t}-\hat{u}_{j,t-1}\right)^2,
\label{eq:ractionrate}
\end{equation}
where $\hat{u}_{j,t}$ is the normalized torque command at time $t$.
This term penalizes rapid torque changes for smooth assistance.

Stages 3 and 4 also include two auxiliary stability penalties: a joint constraint force penalty $r_{\text{constraint}} = -\Delta t \cdot \max_k(|f_k^{\text{joint}}|/mg)$, which penalizes joint limit forces normalized by body weight $mg$; and a foot contact force penalty $r_{\text{foot}} = -\Delta t \cdot \max(0,\,(|f_r|+|f_l|)/mg - 1.2)$, which penalizes total foot contact forces exceeding 1.2 times body weight; both are inactive in Stages~1 and~2, where imitation rewards suffice to enforce stable gait. Stage~4 produces the final co-adapted policies; reward weights for all stages are listed in Table~\ref{tab:reward_weights}.

The framework is modular: the architecture and simulation interface are fixed across stages, and four settings are configurable: (i)~\emph{actor learning activation}: an actor can participate in rollout while its learning signal is disabled, so that only the other actor updates; (ii)~\emph{parameter freezing}: one actor's weights can be fixed while the other updates, decoupling the two adaptation problems; (iii)~\emph{observation augmentation}: an actor's input interface can be extended mid-training by appending new dimensions with existing weights preserved; and (iv)~\emph{reward activation}: stage-specific reward terms can be toggled without changing the network topology or the PPO update.
The hyperparameters for the four-stage curriculum are listed in Table~\ref{tab:ppo_params}. All training procedures were conducted on an Intel Xeon W-2145 (3.70\,GHz, 8-core), achieving approximately 28\,h per 100\,M simulation steps. 

\begin{table}[t]
\centering
\caption{PPO training hyperparameters. Values are shared across stages unless noted otherwise.}
\label{tab:ppo_params}
\setlength{\tabcolsep}{4pt}
\begin{tabular}{lcc}
\toprule
Parameter & S1 & S2--S4 \\
\midrule
Learning rate         & $5{\times}10^{-5}$ & $3{\times}10^{-5}$ \\
PPO clip range        & 0.15   & 0.15 \\
Rollout steps / env   & 2048   & 2048 \\
Minibatch size        & 16384  & 16384 \\
Epochs per update     & 20     & 20 \\
Discount $\gamma$     & 0.99   & 0.99 \\
GAE $\lambda$         & 0.95   & 0.95 \\
Target KL             & 0.01   & 0.01 \\
Max gradient norm     & 0.5    & 0.5 \\
Entropy coefficient   & 0.001  & 0.001 / 0.003$^{a}$ \\
Parallel environments & 32     & 32 \\
\bottomrule
\multicolumn{3}{p{0.95\linewidth}}{\footnotesize
$^{a}$ 0.003 for Stages 3 and 4.}
\end{tabular}
\end{table}

\begin{table}[t]
\caption{Hip exoskeleton hardware specifications.}
\label{tab:exo_hardware}
\centering
\footnotesize
\setlength{\tabcolsep}{4pt}
\begin{tabular}{p{0.42\columnwidth} p{0.48\columnwidth}}
\toprule
\textbf{Item} & \textbf{Specification} \\
\midrule
Configuration & Custom-designed bilateral hip exoskeleton, Fig.~\ref{fig:framework}. \\
Assistance plane & Sagittal plane (hip flex/ext) \\
Active DOF & 1 per side (hip flexion and extension) \\
Passive DOF & 1 per side (hip abd/add) \\
Actuator & MyActuator X8-25 BLDC (2 total) \\
Peak torque capacity & 25~Nm per hip \\
Total mass & 5.94~kg \\
Onboard computer & Raspberry Pi 4B \\
Controller input sensing & Built-in motor encoders \\
Command and logging rate & 50~Hz and 50~Hz \\
Communication & CAN bus \\
Structure materials & Aluminum alloy, carbon tube, 3D printed parts using Nylon-Carbon fiber material (Onyx). \\
\bottomrule
\end{tabular}
\end{table}

\subsection{Hip Exoskeleton Hardware}
\label{subsec:hardware}
A custom bilateral hip exoskeleton provides sagittal-plane hip assistance during treadmill walking (Fig.~\ref{fig:framework}, right). Each side has an actuated hip flexion/extension joint and a passive abduction/adduction joint, delivering assistance while accommodating fitting and natural lateral motion. The exoskeleton hip axis was aligned with the anatomical hip axis during fitting. 

The Stage 4 policy was deployed on a Raspberry Pi 4B on the exoskeleton frame, communicating with the MyActuator X8-25 motors via CAN bus at 50~Hz. It receives joint states from the motor encoders and outputs a normalized torque command in $[-1,1]$, scaled by the preset per-hip torque limit to give the commanded hip torque. 

During treadmill walking, standard safety protections were applied, including access to an emergency stop, software torque limit, and operator supervision. Hardware specifications of the device are summarized in Table~\ref{tab:exo_hardware}.

\subsection{Experimental Design}
\label{subsec:exp_design}
Five participants with no mobility impairment enrolled in the treadmill walking experiment under IRB approval with informed consent. Each subject completed a fixed-order protocol: a 5\,min familiarization, a 5\,min no-assist trial (torque commanded to zero, back-drivable actuators), and two 5\,min assist trials at 10 and 15\,Nm limits, with 2\,min rest between conditions. All trials were performed at 1.25\,m/s (2.8\,mph) on a treadmill. The 25\,Nm training limit is the actuator peak. The human trials used conservative 10 and 15\,Nm limits for safety, and since the policy outputs a normalized command, these exercise a subset of the trained range.

We conducted analyses of bilateral hip joint angle, angular velocity, and commanded torque, along with assistance mechanical power \(P_{\text{command}}=\tau_{\text{command}}\omega\). Angular velocity was filtered with a One Euro filter before power calculation~\cite{casiez20121}, and gait cycles were segmented using peak hip flexion as the reference event. 

\subsection{Exoskeleton Torque and Power Analysis}
To quantify the mechanical output of the exoskeleton controller,
four metrics were computed per subject and condition following~\cite{lim2023parametric}:
root-mean-square torque ($\tau_\text{RMS}$, Nm), mean peak torque
($\tau_\text{MAX}$, Nm), mean positive power (MPP, W), and mean
negative power (MNP, W).
Gait cycles were segmented at peak hip flexion, used as the reference
event in the absence of ground-reaction force sensors.
For each cycle, MPP and MNP were computed as
\begin{equation}
  \text{MPP} = \frac{1}{n}\!\sum_{\tau_i\dot{q}_i>0}\!\tau_i\dot{q}_i,
  \quad
  \text{MNP} = \frac{1}{n}\!\sum_{\tau_i\dot{q}_i<0}\!\tau_i\dot{q}_i,
\end{equation}
where $\tau_i$ and $\dot{q}_i$ are the filtered hip joint torque and
angular velocity, and $n$ is the total number of samples per cycle.
All metrics were averaged across gait cycles per subject.

\section{Results}

\subsection{Simulation Evaluation of Stable Walking and Effects of Assistance}

\begin{figure}[t]
  \centering
  \includegraphics[width=\columnwidth]{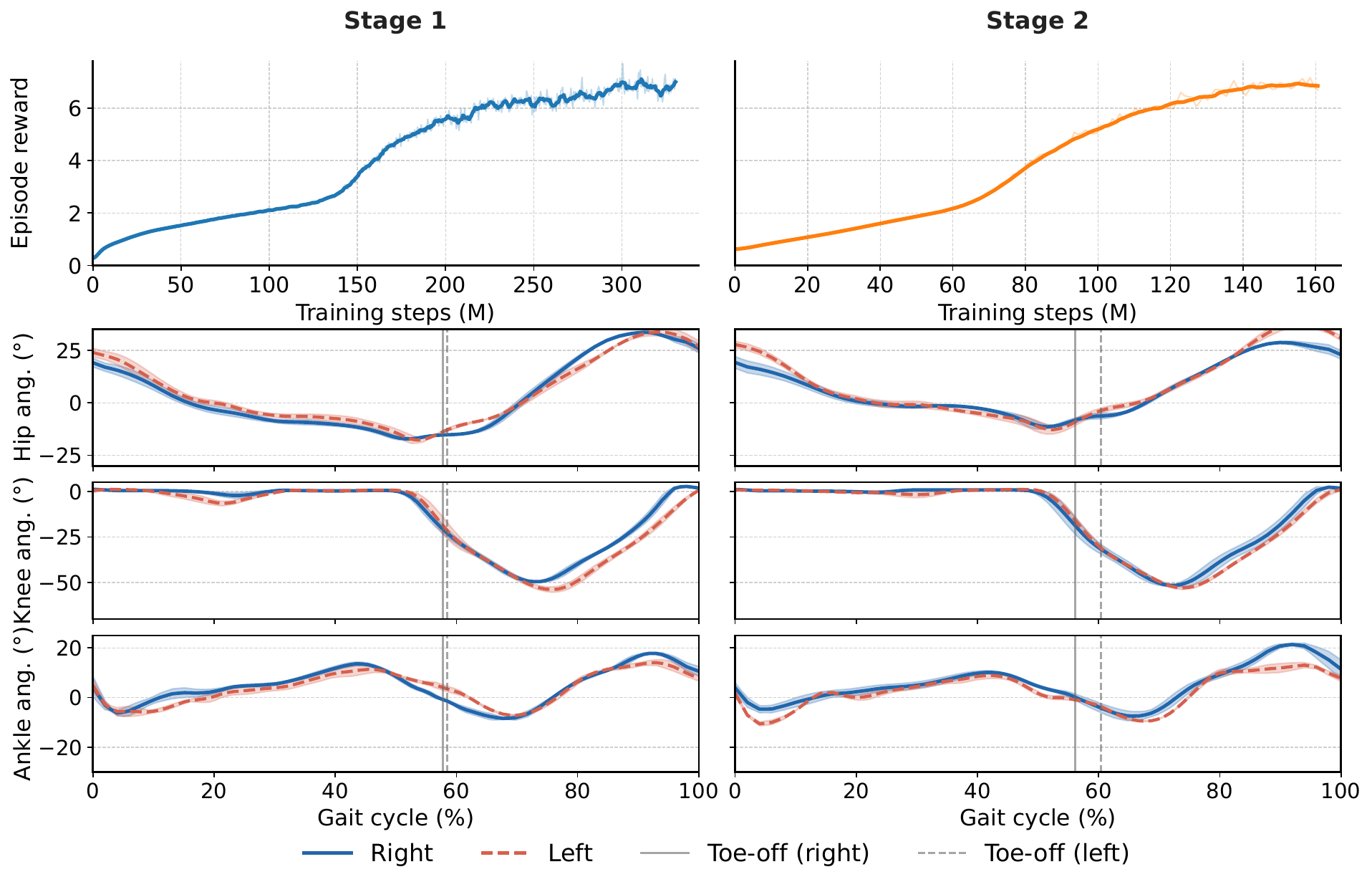}
    \caption{Training reward curves and joint kinematics for Stages~1 and~2 (left and right columns, respectively). \textbf{Top:} Episode reward over training steps. \textbf{Bottom:} Mean $\pm$1\,s.d.\ hip, knee, and ankle angle trajectories over the normalized gait cycle; right limb (solid blue), left limb (dashed red). Vertical lines: mean toe-off.}
  \label{fig:stage12}
\end{figure}

Stage~1 trains the human musculoskeletal actor without any exoskeleton, converging over approximately 320\,M simulation steps (Fig.~\ref{fig:stage12}). The learned gait shows strong bilateral symmetry across all joints (Pearson $r > 0.92$), with toe-off at $57.8 \pm 1.1$\,\% (right) and $58.5 \pm 0.9$\,\% (left) of the gait cycle, consistent with physiological norms for treadmill walking~\cite{perry2024gait}. In Stage~2, the exoskeleton structure is attached with its torque output set to zero, and the human actor is fine-tuned to accommodate the added device mass and inertia. Starting from the Stage~1 policy, convergence is reached within approximately 160\,M steps (Fig.~\ref{fig:stage12}), roughly half the steps required in Stage~1. Joint trajectories are well preserved (RMSE $< 3.6^\circ$ for all joints), and bilateral symmetry is maintained~\cite{forczek2012evaluation}.

\begin{figure}[t]
  \centering
  \includegraphics[width=0.9\columnwidth]{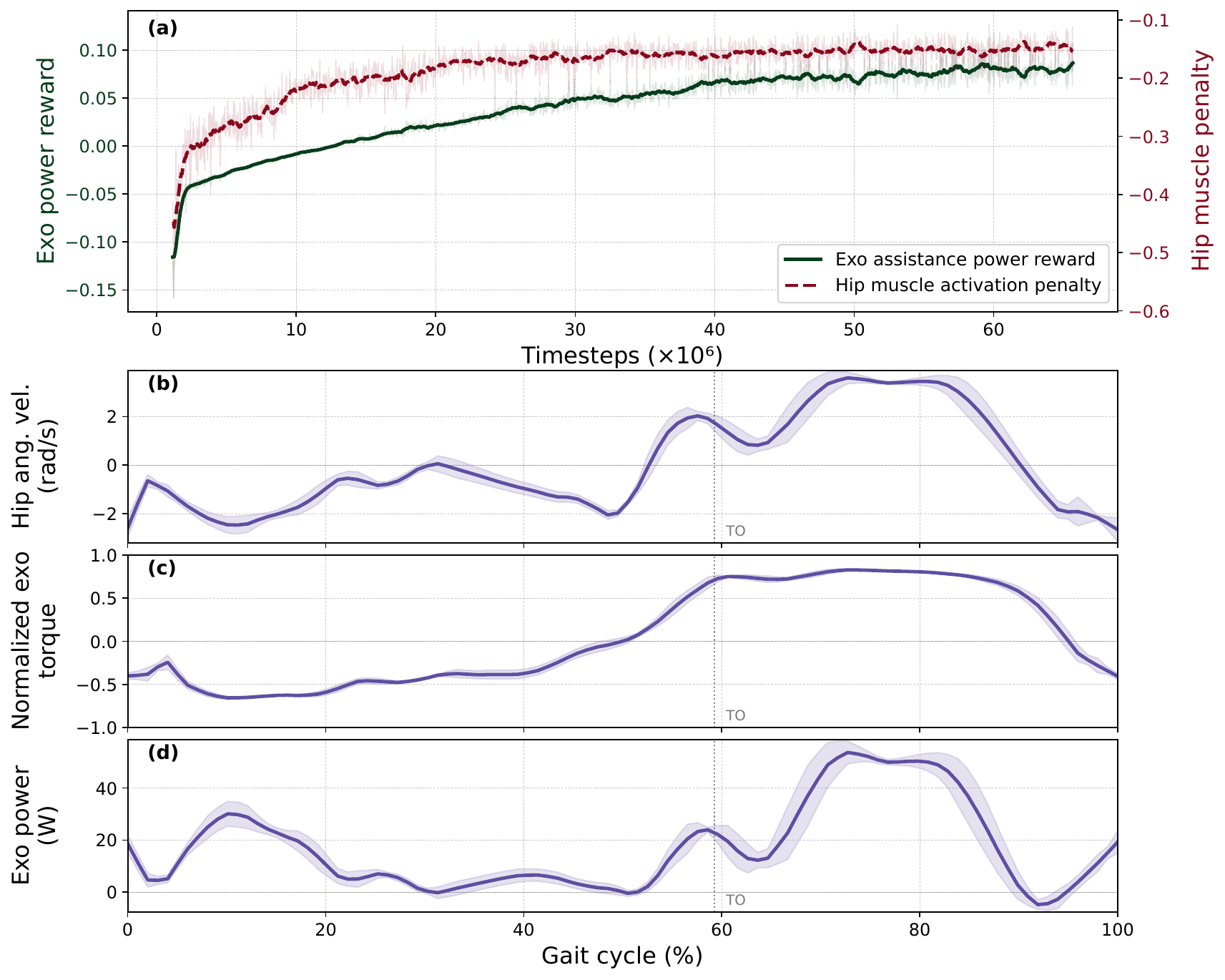}
    \caption{Stage~4 co-adapted profiles. \textbf{(a)}~Exoskeleton assistance power reward (left axis, green) and hip muscle activation penalty (right axis, red dashed) over training. \textbf{(b)}~Hip angular velocity over the gait cycle (positive: flexion; negative: extension). \textbf{(c)}~Normalized exoskeleton torque. \textbf{(d)}~Exoskeleton mechanical power. Panels~(b--d): bilateral average; shaded region: $\pm$1\,SD. Dotted vertical line: mean toe-off.}
  \label{fig:stage34}
\end{figure}

Stage~3 establishes an initial flexion-assist timing pattern in approximately 1.1\,M simulation steps. Stage~4 refines this through full co-adaptation over approximately 66\,M steps.
As shown in Fig.~\ref{fig:stage34}a, the exoskeleton assistance power reward increases from $-0.12$ to $0.08$ and the hip muscle activation penalty decreases in magnitude over the course of training. The final co-adapted torque profile (Fig.~\ref{fig:stage34}c), plotted alongside the hip angular velocity (Fig.~\ref{fig:stage34}b), indicates that the exoskeleton torque is predominantly aligned with joint motion throughout the gait cycle, with peak normalized torque of 0.84 occurring at 75\,\% of the gait cycle during the late-swing flexion phase, reducing the negative-work time fraction (fraction of the gait cycle where $P = \tau\dot{q} < 0$) to 10\,\%, as reflected in Fig.~\ref{fig:stage34}d.

\begin{figure}[t]
  \centering
  \includegraphics[width=0.9\columnwidth]{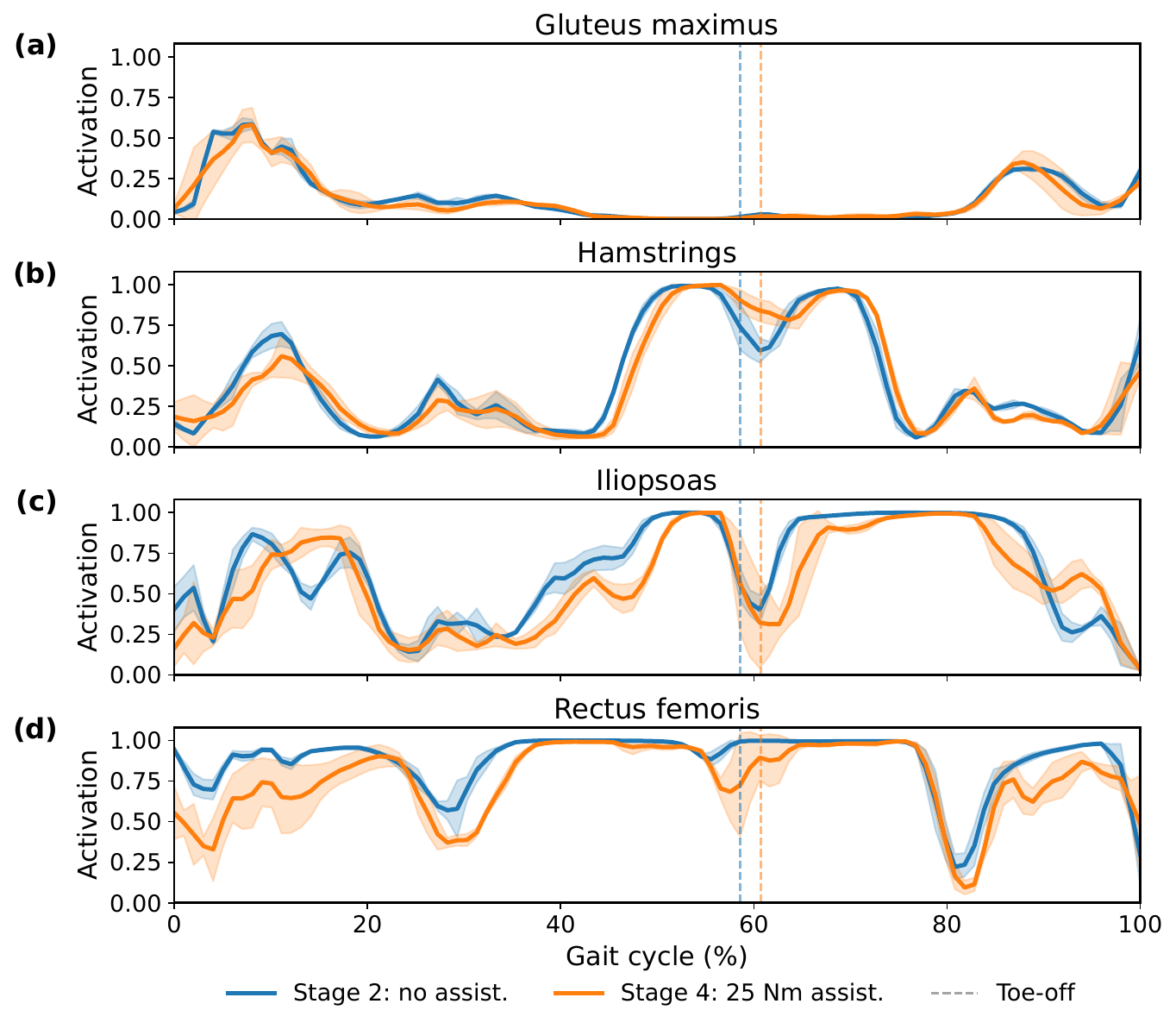}
  \caption{Right-side hip muscle activation over the gait cycle comparing Stage~2 (no assistive torque, blue) and Stage~4 (maximum assistive torque 25\,Nm, orange). Shaded bands: mean $\pm$ 1\,s.d. Dashed vertical lines mark mean toe-off for each condition.}
  \label{fig:muscle}
\end{figure}

The reduction in hip muscle activation is further quantified in Fig.~\ref{fig:muscle}, comparing right-side activation between Stage~2 (no assistive torque) and Stage~4 (maximum assistive torque, $\tau_{\max} = 25$\,Nm). All four muscles show reduced mean activation with exoskeleton assistance, averaging 10.1\,\% across muscles. The largest reductions are in the primary hip flexors: rectus femoris ($-$13.5\,\%) and iliopsoas ($-$10.5\,\%), whose swing-phase activation is partially substituted by the exoskeleton's flexion torque. Hamstrings decrease by 9.7\,\% and gluteus maximus by 6.6\,\%, with the smaller extensor reductions consistent with the learned torque profile providing less direct offloading during stance.

\begin{figure}[t]
  \centering
  \includegraphics[width=0.9\columnwidth]{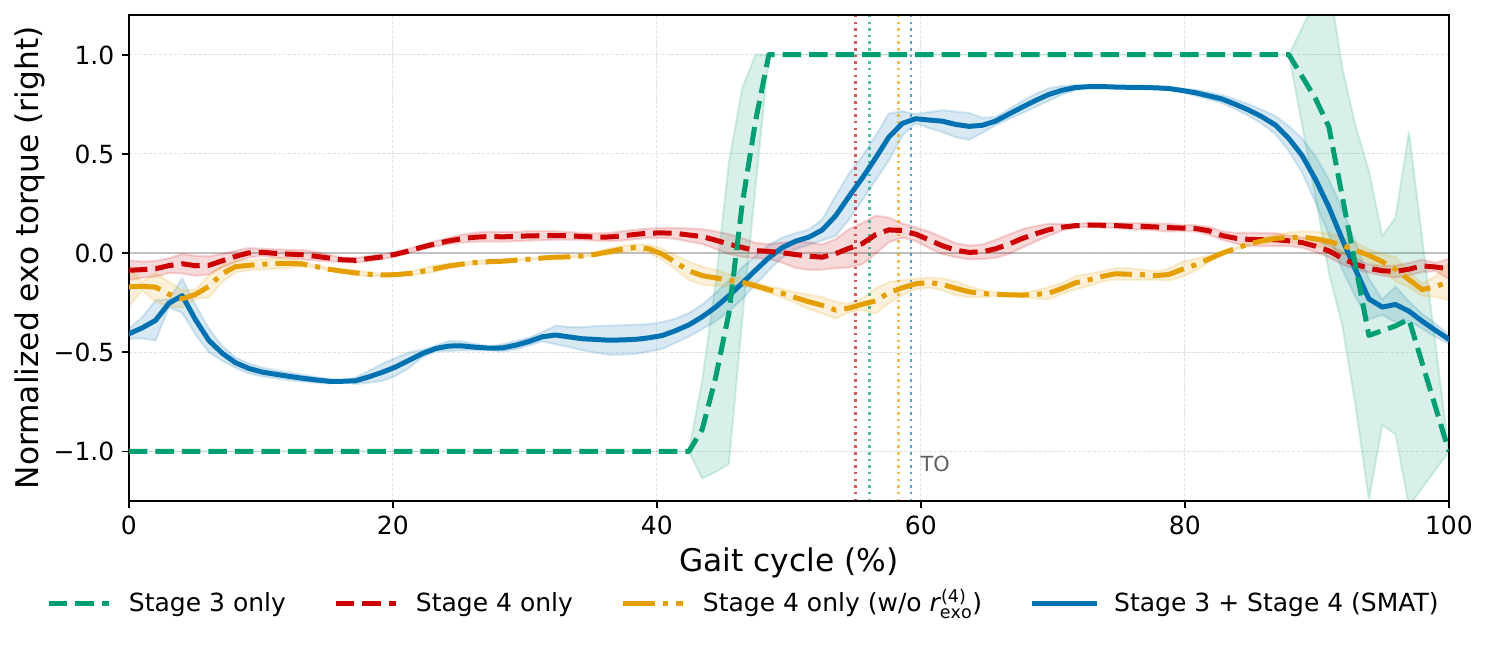}
    \caption{Ablation comparing right-hip normalized exo torque over the gait cycle
    under four conditions: \textbf{Stage~3 only} without Stage~4 co-adaptation
    (dashed green); \textbf{Stage~4 only} initialized from Stage~2 without
    Stage~3 pre-training (dashed red, mean across three runs);
    \textbf{Stage~4 only without $r_\mathrm{exo}^{(4)}$}, initialized from
    Stage~2 (dash-dot orange); and the full \textbf{Stage~3 + Stage~4 (SMAT)}
    pipeline (solid blue). Dotted vertical lines mark mean toe-off.}
  \label{fig:ablation}
\end{figure}

\begin{table*}[t]
\caption{Per-Subject Exoskeleton Joint Torque and Power Metrics Across Assistance Conditions}
\label{tab:torque_power}
\centering
\renewcommand{\arraystretch}{1.05}
\begin{tabular}{l|cccc|cccc}
\hline
 & \multicolumn{4}{c}{10\,Nm} & \multicolumn{4}{c}{15\,Nm} \\
\cline{2-5}\cline{6-9}
Subject No. & $\tau_\text{RMS}$ (Nm) & $\tau_\text{MAX}$ (Nm) & MPP (W) & MNP (W) & $\tau_\text{RMS}$ (Nm) & $\tau_\text{MAX}$ (Nm) & MPP (W) & MNP (W) \\
\hline
S1 & 6.12 & 8.12 & 14.93 & $-$0.16 & 9.40 & 12.25 & 26.58 & $-$0.14 \\
S2 & 6.13 & 8.14 & 17.26 & $-$0.11 & 9.67 & 12.13 & 29.91 & $-$0.10 \\
S3 & 5.87 & 8.07 & 11.74 & $-$0.15 & 9.07 & 12.39 & 20.02 & $-$0.18 \\
S4 & 5.89 & 7.95 & 12.90 & $-$0.09 & 9.13 & 12.03 & 22.01 & $-$0.18 \\
S5 & 5.97 & 8.02 & 11.27 & $-$0.10 & 9.14 & 12.30 & 20.51 & $-$0.10 \\
\hline
Mean ($\pm$SD) & $6.0\pm0.1$ & $8.1\pm0.1$ & $13.6\pm2.5$ & $-0.1\pm0.0$ & $9.3\pm0.3$ & $12.2\pm0.1$ & $23.8\pm4.3$ & $-0.1\pm0.0$ \\
\hline
\multicolumn{9}{l}{\footnotesize $\tau_\text{RMS}$: root-mean-square torque;\quad $\tau_\text{MAX}$: mean peak torque;\quad MPP: mean positive power;\quad MNP: mean negative power.} \\
\multicolumn{9}{l}{\footnotesize Gait cycles segmented at peak hip flexion angle. All metrics computed for the right leg.} \\
\end{tabular}
\end{table*}

\textit{Ablation: necessity of Stage~3 and Stage~4.}
As shown in Fig.~\ref{fig:ablation}, three ablation conditions reveal the necessity of both stages. Without Stage~4 co-adaptation (Stage~3 only), the torque saturates at $\pm\hat{\tau}_{\max}$ for most of the gait cycle and reverses abruptly near toe-off, imposing impulsive hip loading that would pose safety concerns in physical human--robot interaction~\cite{moreno2009analysis}. Without Stage~3 pre-training (\textbf{Stage~4 only}), training collapses to a local optimum: the second term of~\eqref{eq:rexoreward_stage4} rewards minimizing torque magnitude, so the policy discovers that producing no torque maximizes that reward component, yielding near-zero output (peak normalized torque 0.14 vs.\ 0.84, an 83\,\% reduction), with 32\,\% of this minimal torque opposing joint motion (vs.\ 10\,\% in the full SMAT pipeline). Removing $r_\mathrm{exo}^{(4)}$ entirely (\textbf{Stage~4 only w/o $r_\mathrm{exo}^{(4)}$}) causes the exoskeleton to converge to a slight constant extension torque, which persistently opposes hip flexion during swing and results in negative exo power for 40\,\% of the gait cycle. These three conditions confirm that both Stage~3 pre-training and Stage~4 co-adaptation are required components of the full SMAT pipeline.

\subsection{Verification of Assistance Efficiency and Speed Generalization with Open Source Dataset}

\begin{figure}[t]
  \centering
  \includegraphics[width=0.9\columnwidth]{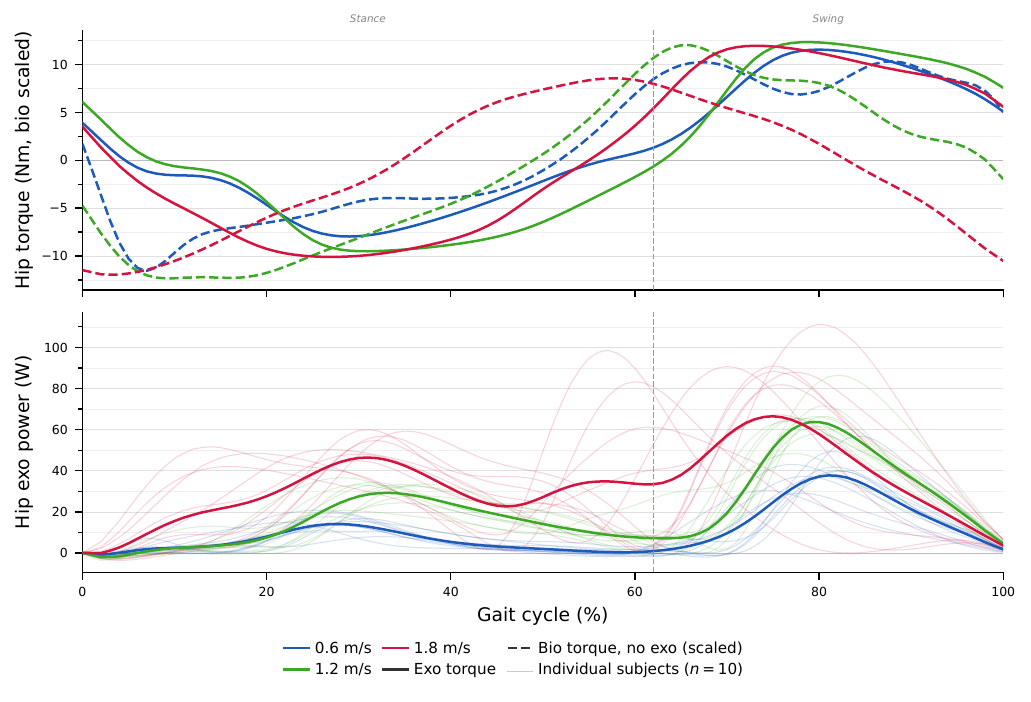}
  \caption{Speed generalization on the GT dataset~\cite{molinaro2024task} (10 subjects,
  0.6, 1.2, and 1.8\,m/s; blue/green/red).
  \emph{Top:} Gait-cycle-normalized group mean hip torque; solid: exo torque,
  dashed: biological hip torque (no-exo condition, amplitude scaled for timing comparison).
  \emph{Bottom:} Hip exo power; thin lines: individual subjects, bold: group mean.
  Dashed vertical line: toe-off ($\approx$62\,\%).}
  \label{fig:gt_validation}
\end{figure}

To evaluate generalizability beyond training conditions, we assessed the SMAT-trained exo actor on an open source dataset~\cite{molinaro2024task}, comprising 10 subjects walking at 0.6, 1.2, and 1.8\,m/s. Importantly, the policy receives only hip kinematics as input, enabling potential generalization to different walking speeds without explicit speed-based adjustment.

Fig.~\ref{fig:gt_validation} shows gait-cycle-normalized exo torque profiles alongside the biological hip torque (amplitude scaled to match exo peak for timing comparison). Peak assistive torque remained consistent across all speeds (11.6--12.3\,Nm), well within the 15\,Nm chosen maximum. The exo torque profile lagged biological hip torque by 9--20\,\% of the gait cycle (${\sim}$156--210\,ms, by cross-correlation of torque waveforms). In the work by Molinaro et al.~\cite{molinaro2024task}, the authors intentionally imposed a 125\,ms hip-assistance delay to maximize positive mechanical work. Our policy produces a similar phase offset through reward optimization alone, without explicit timing constraints~\cite{kang2019real}. Peak exo power reached 38--67\,W in early-to-mid swing, consistent with the biological assistive demand, and timing remained stable across all speeds and subjects without subject-specific retraining.

\begin{figure}[t]
  \centering
  \includegraphics[width=0.90\columnwidth]{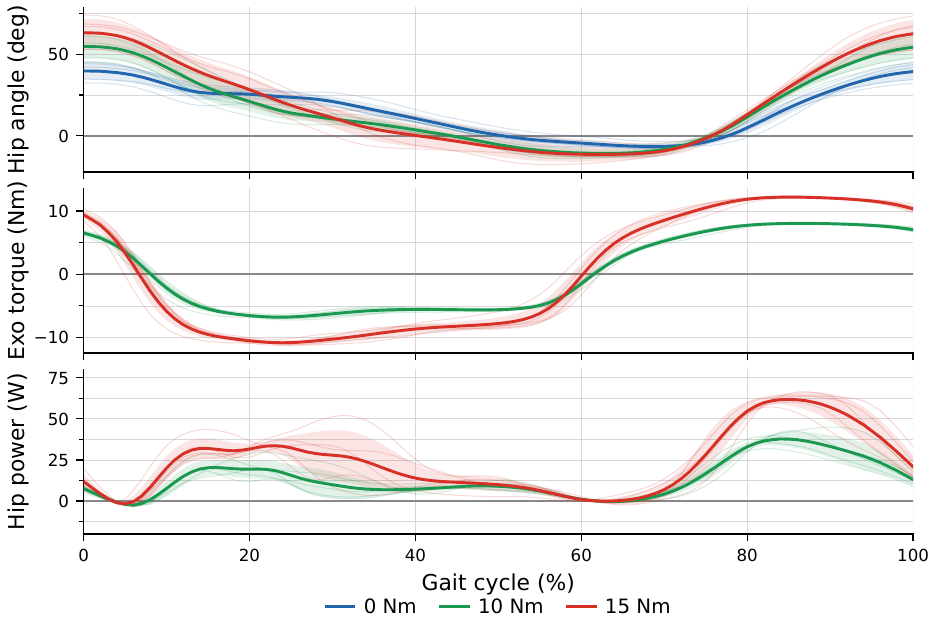}
    \caption{Hip kinematics, exo torque, and mechanical power under assisted conditions (0\,\% = peak hip flexion; 0\,Nm: blue, 10\,Nm: green, 15\,Nm: red). \textbf{Top:} Group-mean right hip angle. \textbf{Middle:} Exoskeleton torque. \textbf{Bottom:} Right hip assistance power. Shaded bands: $\pm1$\,SD; thin lines: subject means.}
  \label{fig:kinematics}
\end{figure}

\subsection{Validation with Human Experiment}

We evaluated the sim-to-real transfer of the SMAT-trained policy on five healthy participants (S1--S5) across three assistance levels: 0\,Nm, 10\,Nm, and 15\,Nm. Fig.~\ref{fig:kinematics} and Table~\ref{tab:torque_power} summarize the measured hip kinematics and joint-level mechanical output. \textit{Hip kinematics.} In Fig.~\ref{fig:kinematics} (top), the waveform shape remained consistent across conditions: peak hip extension occurred in late stance ($\approx$62--68\,\% gait cycle) and hip flexion accelerated into swing phase, matching the normal gait pattern. This preserved timing indicates that assistance is modulated without altering gait timing.

\textit{Joint work analysis.} All metrics are reported for the right leg. Fig.~\ref{fig:kinematics} (middle) shows that the exo torque waveform is consistent across subjects, with a positive assistive peak in mid-swing. Peak torque scales with the commanded level ($8.1\pm0.1$\,Nm at 10\,Nm limit, $12.2\pm0.1$\,Nm at 15\,Nm limit). As shown in Fig.~\ref{fig:kinematics} (bottom) and reported in Table~\ref{tab:torque_power}, MPP increased from $13.6\pm2.5$\,W at 10\,Nm to $23.8\pm4.3$\,W at 15\,Nm, confirming that the policy scales mechanical energy delivery in proportion to the commanded assistance level, while MNP remained very small (${\approx}{-0.1}$\,W) in both conditions, indicating low resistive losses. Subject consistency confirms that the policy delivers assistance at the correct timing without subject-specific parameter adjustment. At a comparable $\tau_\text{RMS}$ of approximately 6\,Nm, the delayed output feedback controller of Lim et al.~\cite{lim2019delayed} reported an MPP of 9.9--11.6\,W during treadmill walking at 4\,km/h, while our policy achieved 13.6\,W at the same RMS torque, indicating higher positive-power delivery per unit RMS torque than this reference controller.

\section{Discussion}

The ablation clarifies the roles of Stages 3 and 4. Without Stage~3, Stage~4 begins from a random exoskeleton policy and converges to a near-zero torque local optimum, whether or not $r_{\text{exo}}^{(4)}$ is included, in which the exoskeleton avoids assistance to minimize disturbance to the imitation and muscle penalties. Stage 3 prevents this collapse by isolating assistance timing learning under controlled conditions. Freezing the human policy removes non-stationarity, while the reduced torque limit ($\tau_{\max}=6$ Nm) constrains destabilizing exploration. The additional reward $r_{\text{exo}}^{(3)}$ explicitly encourages torque aligned with joint velocity, establishing a positive mechanical power pattern before full co-adaptation. This warm-start ensures that Stage 4 begins from a meaningful assistance strategy rather than from zero-output behavior.

From a sim-to-real perspective, sensing modality introduces practical limitations. In simulation, hip angles correspond to anatomical joint motion, whereas on hardware the policy relies on exoskeleton encoder measurements. Due to soft-tissue compliance and strap looseness, encoder-derived angles may overestimate true hip motion and yield noisier velocities than motion capture, degrading timing precision. Training directly on encoder-only kinematics, or replacing encoders with thigh- and pelvis-mounted IMUs~\cite{seel2014imu}, could improve phase estimation and alignment.

The learned controller operates below the torque limit. This behavior is shaped by the Stage 4 saturation penalty (Eq.~\eqref{eq:rexoreward_stage4}, $\delta=0.8$), which discourages impulsive loading, eliminates high-amplitude torque bursts, and prevents resistive extension strategies that push the negative-work fraction higher. We also observed mild inter-limb asymmetry in hip angular velocity, likely due to the absence of an explicit symmetry constraint during training. Incorporating a bilateral symmetry term could further regularize assistance profiles.

In simulation, hip muscle activation decreased by 10.1\% on average. However, physiological validation remains necessary. Real users may redistribute effort across muscle groups or gradually alter coordination patterns as adaptation progresses. Future studies incorporating EMG and indirect calorimetry \cite{sylos2014emg, seo2016fully} will determine whether simulated offloading translates to measurable metabolic benefit. The current evaluation included five healthy participants at a single treadmill speed. Broader validation across larger cohorts, varied speeds, slopes, and perturbations, clinical populations, and deployment outside the laboratory will better characterize generalization \cite{chen2024learning}. Two points about the method remain open. First, the staged design is inspired by, not a validated model of, human motor adaptation. Second, our ablation shows that staging reliably avoids the non-assistive local optimum under our reward design. A broader comparison against single-stage co-training~\cite{luo2024experiment} across reward settings is left to future work.

\section{Conclusions}

Jointly training human and exoskeleton policies is non-stationary and prone to poor local optima. SMAT addresses this with four sequential stages, gait learning, passive adaptation, assistance-timing pre-training, and co-adaptation, each shown necessary by the ablation. The final trained controller reduced hip muscle activation by 10.1\,\% in simulation, generalized to unseen subjects across 0.6--1.8\,m/s without explicit speed adjustment, and transferred to hardware without subject-specific retraining. The deployed controller delivered consistent assistance timing and predominantly positive mechanical power (MPP: 13.6--23.8\,W, with negative power below 1.3\,\% of MPP) across five participants. Relative to a representative delayed output feedback controller~\cite{lim2019delayed}, the proposed controller achieves more effective positive-power delivery per unit RMS torque, indicating improved assistance efficiency. Notably, the policy yields an emergent assistance-phase delay of 9--20\,\% of the gait cycle, qualitatively consistent with the deliberately delayed hip-assistance phasing used to increase positive mechanical work in~\cite{molinaro2024task}.

\bibliographystyle{IEEEtran}
\bibliography{references}

@article{luo2023robust,
  title={Robust walking control of a lower limb rehabilitation exoskeleton coupled with a musculoskeletal model via deep reinforcement learning},
  author={Luo, Shuzhen and Androwis, Ghaith and Adamovich, Sergei and Nunez, Erick and Su, Hao and Zhou, Xianlian},
  journal={Journal of neuroengineering and rehabilitation},
  volume={20},
  number={1},
  pages={34},
  year={2023},
  publisher={Springer}
}

@article{luo2024experiment,
  title={Experiment-free exoskeleton assistance via learning in simulation},
  author={Luo, Shuzhen and Jiang, Menghan and Zhang, Sainan and Zhu, Junxi and Yu, Shuangyue and Dominguez Silva, Israel and Wang, Tian and Rouse, Elliott and Zhou, Bolei and Yuk, Hyunwoo and others},
  journal={Nature},
  volume={630},
  number={8016},
  pages={353--359},
  year={2024},
  publisher={Nature Publishing Group UK London}
}

@inproceedings{tan2025myoassist,
  title={Myoassist 0.1: Myosuite for dexterity and agility in bionic humans},
  author={Tan, Chun Kwang and Wang, Cheryl and Lyu, Shirui and Hodossy, Balint K and Schumacher, Pierre and Wilson, Elizabeth B and Caggiano, Vittorio and Kumar, Vikash and Farina, Dario and Gionfrida, Letizia and others},
  booktitle={2025 International Conference On Rehabilitation Robotics (ICORR)},
  pages={437--442},
  year={2025},
  organization={IEEE}
}

@article{caggiano2022myosuite,
  title={MyoSuite--A contact-rich simulation suite for musculoskeletal motor control},
  author={Caggiano, Vittorio and Wang, Huawei and Durandau, Guillaume and Sartori, Massimo and Kumar, Vikash},
  journal={arXiv preprint arXiv:2205.13600},
  year={2022}
}

@article{chiappa2024acquiring,
  title={Acquiring musculoskeletal skills with curriculum-based reinforcement learning},
  author={Chiappa, Alberto Silvio and Tano, Pablo and Patel, Nisheet and Ingster, Abiga{\"\i}l and Pouget, Alexandre and Mathis, Alexander},
  journal={Neuron},
  volume={112},
  number={23},
  pages={3969--3983},
  year={2024},
  publisher={Elsevier}
}

@article{molinaro2024task,
  title={Task-agnostic exoskeleton control via biological joint moment estimation},
  author={Molinaro, Dean D and Scherpereel, Keaton L and Schonhaut, Ethan B and Evangelopoulos, Georgios and Shepherd, Max K and Young, Aaron J},
  journal={Nature},
  volume={635},
  number={8038},
  pages={337--344},
  year={2024},
  publisher={Nature Publishing Group UK London}
}

@article{lim2023parametric,
  title={Parametric delayed output feedback control for versatile human-exoskeleton interactions during walking and running},
  author={Lim, Bokman and Choi, Byungjune and Roh, Changhyun and Hyung, Seungyong and Kim, Yong-Jae and Lee, Younbaek},
  journal={IEEE Robotics and Automation Letters},
  volume={8},
  number={8},
  pages={4497--4504},
  year={2023},
  publisher={IEEE}
}

@article{leem2026exo,
  title={Exo-Plore: Exploring Exoskeleton Control Space through Human-aligned Simulation},
  author={Leem, Geonho and Lee, Jaedong and Lee, Jehee and Song, Seungmoon and Won, Jungdam},
  journal={arXiv preprint arXiv:2601.22550},
  year={2026}
}

@article{yuan2026gait,
  title={Gait Asymmetry from Unilateral Weakness and Improvement With Ankle Assistance: a Reinforcement Learning based Simulation Study},
  author={Yuan, Yifei and Androwis, Ghaith and Zhou, Xianlian},
  journal={arXiv preprint arXiv:2602.18862},
  year={2026}
}

@article{ratnakumar2026reinforcement,
  title={Reinforcement-Learning-Based Assistance Reduces Squat Effort with a Modular Hip--Knee Exoskeleton},
  author={Ratnakumar, Neethan and Tohfafarosh, Mariya Huzaifa and Jauhri, Saanya and Zhou, Xianlian},
  journal={arXiv preprint arXiv:2602.17794},
  year={2026}
}

@article{lim2019delayed,
  title={Delayed output feedback control for gait assistance with a robotic hip exoskeleton},
  author={Lim, Bokman and Lee, Jusuk and Jang, Junwon and Kim, Kyungrock and Park, Young Jin and Seo, Keehong and Shim, Youngbo},
  journal={IEEE Transactions on Robotics},
  volume={35},
  number={4},
  pages={1055--1062},
  year={2019},
  publisher={IEEE}
}

@article{schulman2017proximal,
  title={Proximal policy optimization algorithms},
  author={Schulman, John and Wolski, Filip and Dhariwal, Prafulla and Radford, Alec and Klimov, Oleg},
  journal={arXiv preprint arXiv:1707.06347},
  year={2017}
}

@article{rodriguez2021systematic,
  title={Systematic review on wearable lower-limb exoskeletons for gait training in neuromuscular impairments},
  author={Rodr{\'\i}guez-Fern{\'a}ndez, Antonio and Lobo-Prat, Joan and Font-Llagunes, Josep M},
  journal={Journal of neuroengineering and rehabilitation},
  volume={18},
  number={1},
  pages={22},
  year={2021},
  publisher={Springer}
}

@article{poggensee2021adaptation,
  title={How adaptation, training, and customization contribute to benefits from exoskeleton assistance},
  author={Poggensee, Katherine L and Collins, Steven H},
  journal={Science Robotics},
  volume={6},
  number={58},
  pages={eabf1078},
  year={2021},
  publisher={American Association for the Advancement of Science}
}

@inproceedings{casiez20121,
  title={1€ filter: a simple speed-based low-pass filter for noisy input in interactive systems},
  author={Casiez, G{\'e}ry and Roussel, Nicolas and Vogel, Daniel},
  booktitle={Proceedings of the SIGCHI Conference on human factors in computing systems},
  pages={2527--2530},
  year={2012}
}

@book{perry2024gait,
  title={Gait analysis: normal and pathological function},
  author={Perry, Jacquelin and Burnfield, Judith},
  year={2024},
  publisher={CRC Press}
}

@inproceedings{bengio2009curriculum,
  title={Curriculum learning},
  author={Bengio, Yoshua and Louradour, J{\'e}r{\^o}me and Collobert, Ronan and Weston, Jason},
  booktitle={Proceedings of the 26th annual international conference on machine learning},
  pages={41--48},
  year={2009}
}

@article{baud2021review,
  title={Review of control strategies for lower-limb exoskeletons to assist gait},
  author={Baud, Romain and Manzoori, Ali Reza and Ijspeert, Auke and Bouri, Mohamed},
  journal={Journal of neuroengineering and rehabilitation},
  volume={18},
  number={1},
  pages={119},
  year={2021},
  publisher={Springer}
}

@article{forczek2012evaluation,
  title={An evaluation of symmetry in the lower limb joints during the able-bodied gait of women and men},
  author={Forczek, Wanda and Staszkiewicz, Robert},
  journal={Journal of human kinetics},
  volume={35},
  pages={47},
  year={2012}
}

@article{kang2019real,
  title={Real-time neural network-based gait phase estimation using a robotic hip exoskeleton},
  author={Kang, Inseung and Kunapuli, Pratik and Young, Aaron J},
  journal={IEEE Transactions on Medical Robotics and Bionics},
  volume={2},
  number={1},
  pages={28--37},
  year={2019},
  publisher={IEEE}
}

@article{seel2014imu,
  title={IMU-based joint angle measurement for gait analysis},
  author={Seel, Thomas and Raisch, Jorg and Schauer, Thomas},
  journal={Sensors},
  volume={14},
  number={4},
  pages={6891--6909},
  year={2014},
  publisher={MDPI}
}

@inproceedings{seo2016fully,
  title={Fully autonomous hip exoskeleton saves metabolic cost of walking},
  author={Seo, Keehong and Lee, Jusuk and Lee, Younbaek and Ha, Taesin and Shim, Youngbo},
  booktitle={2016 IEEE International Conference on Robotics and Automation (ICRA)},
  pages={4628--4635},
  year={2016},
  organization={IEEE}
}

@article{sylos2014emg,
  title={EMG patterns during assisted walking in the exoskeleton},
  author={Sylos-Labini, Francesca and La Scaleia, Valentina and d'Avella, Andrea and Pisotta, Iolanda and Tamburella, Federica and Scivoletto, Giorgio and Molinari, Marco and Wang, Shiqian and Wang, Letian and Van Asseldonk, Edwin and others},
  journal={Frontiers in human neuroscience},
  volume={8},
  pages={423},
  year={2014},
  publisher={Frontiers Media SA}
}

@article{chen2024learning,
  title={Learning to assist different wearers in multitasks: efficient and individualized human-in-the-loop adaptation framework for lower-limb exoskeleton},
  author={Chen, Yu and Miao, Shu and Chen, Gong and Ye, Jing and Fu, Chenglong and Liang, Bin and Song, Shiji and Li, Xiang},
  journal={IEEE Transactions on Robotics},
  volume={40},
  pages={4699--4718},
  year={2024},
  publisher={IEEE}
}

@article{moreno2009analysis,
  title={Analysis of the Human Interaction with a Wearable Lower-Limb Exoskeleton},
  author={Moreno, Juan C and Brunetti, Fernando and Navarro, Enrique and Forner-Cordero, Arturo and Pons, Jos{\'e} L},
  journal={Applied Bionics and Biomechanics},
  volume={6},
  number={2},
  pages={245--256},
  year={2009},
  publisher={Wiley Online Library}
}

\end{document}